\documentclass[10pt,letterpaper]{article}

\usepackage{cogsci}

\cogscifinalcopy

\usepackage[
  style=apa,
  natbib=true,
  annotation=false,
]{biblatex}
\usepackage{authblk}
\usepackage{float}
\usepackage{graphicx}
\usepackage{amsmath}
\usepackage{booktabs}
\usepackage{caption}
\usepackage{subcaption}
\usepackage{tikz}
\usetikzlibrary{shapes, arrows.meta, positioning, calc, shadows.blur, fit}

\definecolor{myblue}{RGB}{70, 130, 180}
\definecolor{myorange}{RGB}{255, 140, 0}
\definecolor{mygray}{RGB}{105, 105, 105}
\definecolor{bgblue}{RGB}{240, 248, 255}
\definecolor{bgorange}{RGB}{255, 245, 238}

\usepackage[most]{tcolorbox}
\usepackage{lipsum} 
\usepackage{hyperref}
\newtcolorbox{hypothesis}[1][]{
  colback=blue!5!white, 
  colframe=blue!75!black, 
  fonttitle=\bfseries,
  title=Hypothesis,
  left=0cm, 
  sharp corners,
  boxrule=0.5mm,
  fontupper=\small,
  #1
}

\graphicspath{{images/}}

\title{Learning to Look before Learning to Like: \\ Incorporating Human Visual Cognition into Aesthetic Quality Assessment}

\author[1]{\mbox{Liwen Yu}}
\author[1,$\dag$]{\mbox{Chi Liu} (chiliu@cityu.edu.mo)}
\author[2]{\mbox{Xiaotong Han}}
\author[1]{\mbox{Congcong Zhu}}
\author[1]{\mbox{Minghao Wang}}
\author[3]{\mbox{Sheng Shen}}
\affil[1]{Faculty of Data Science, City University of Macau, Taipa, Macao, China}
\affil[2]{Zhongshan Ophthalmic Center, Sun Yat-sen University, Guangzhou, China}
\affil[3]{School of computer science, Torrens University Australia, Melbourne, Victoria, Australia}
\affil[$\dag$]{\textit{Corresponding Author}}

\begin{document}

\maketitle

\begin{abstract}
Automated Aesthetic Quality Assessment (AQA) treats images primarily as static pixel vectors, aligning predictions with human-rating scores largely through semantic perception. However, this paradigm diverges from human aesthetic cognition, which arises from dynamic visual exploration shaped by scanning paths, processing fluency, and the interplay between bottom-up salience and top-down intention. We introduce AestheticNet, a novel cognitive-inspired AQA paradigm that integrates human-like visual cognition and semantic perception with a two-pathway architecture. The visual attention pathway, implemented as a gaze-aligned visual encoder (GAVE) pre-trained offline on eye-tracking data using resource-efficient contrast gaze alignment, models attention from human vision system. This pathway augments the semantic pathway, which uses a fixed semantic encoder such as CLIP, through cross-attention fusion. Visual attention provides a cognitive prior reflecting foreground/background structure, color cascade, brightness, and lighting, all of which are determinants of aesthetic perception beyond semantics. Experiments validated by hypothesis testing show a consistent improvement over the semantic-alone baselines, and demonstrate the gaze module as a model-agnostic corrector compatible with diverse AQA backbones, supporting the necessity and modularity of human-like visual cognition for AQA. Our code is available at \href{https://github.com/keepgallop/AestheticNet}{github.com/keepgallop/AestheticNet}

\end{abstract}

\section{Introduction}

\begin{figure}[t!]
\begin{center}
\includegraphics[width=\linewidth]{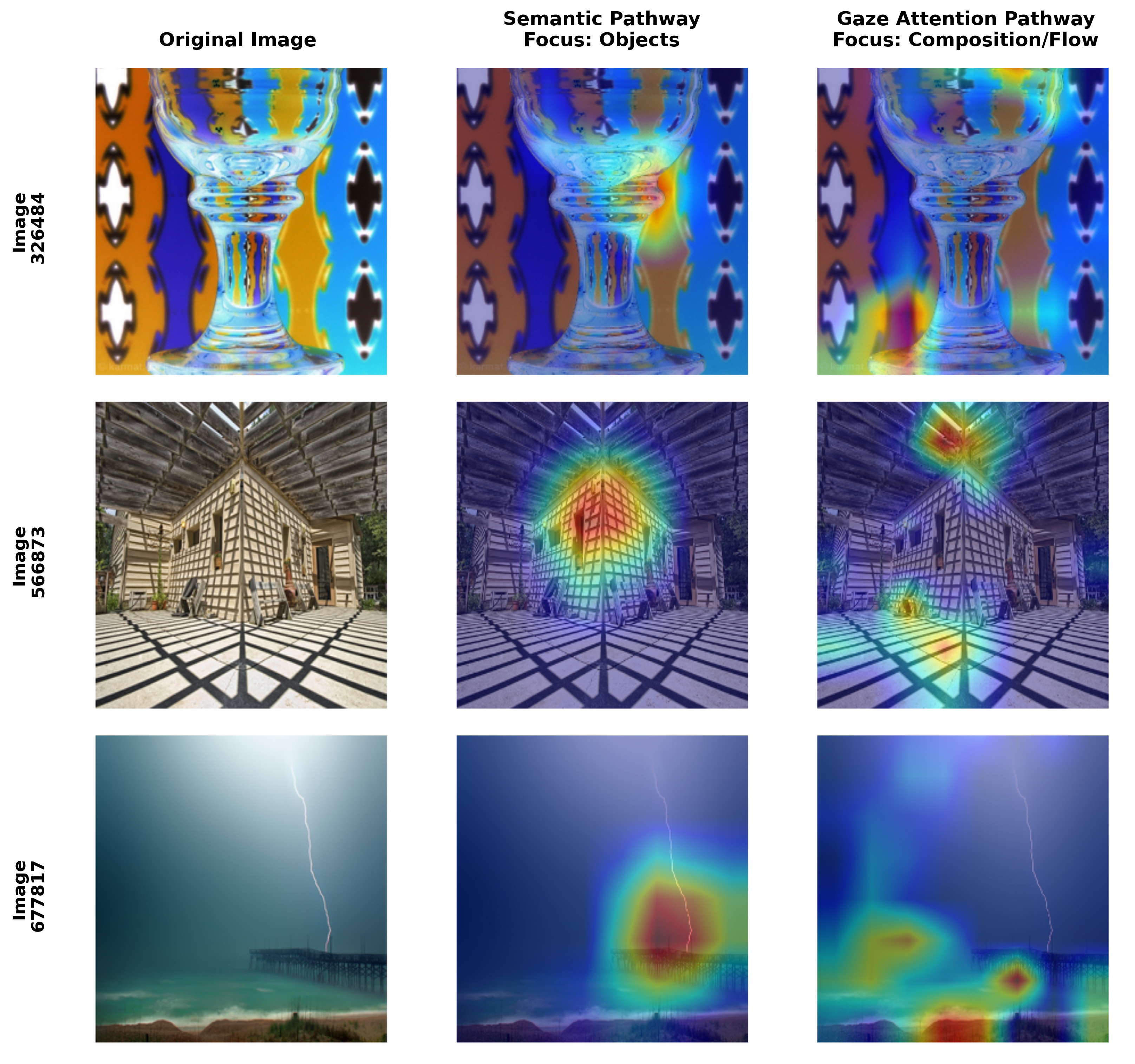}
\end{center}
\caption{Attention divergence between two AQA pathways. The semantic pathway (middle) exhibits object-centric bias, while the gaze attention pathway (right) prioritizes scene compositional properties and visual flow, implying the synergy and complementary effect.}
\label{fig:gradcam}
\end{figure}
Automated Aesthetic Quality Assessment (AQA) is a fundamental computer vision task that facilitates human-computer interaction, image understanding, AI cognition and game design.  The standard paradigm trains a computational model to predict the aesthetic score of an image to match human ratings. Existing models, from CNN-based architectures like NIMA \parencite{talebi2018nima} to recent Large Vision-Language Models like Q-Align \parencite{wu2024qalign}, treat images as static pixel vectors and evaluate aesthetic quality by capturing pixel correlation and semantic content at only a single glance. 

However, despite using human labels as ground truth, this paradigm misaligns with the human visual system (HVS), deviating from the true nature of human aesthetic cognition which is neither instantaneous nor static. Early eye-tracking studies \parencite{yarbus1967eye} have proven that humans actively perceive scenes via scanpaths \parencite{noton1971scanpaths}, driven by a dynamic interplay between bottom-up saliency (e.g., contrast, orientation) and top-down semantic intent. This active exploration for ``looking'' strongly correlates with the outcome of ``liking'' \parencite{shimamura2012aesthetic}. Processing Fluency Theory further suggests that aesthetic pleasure arises from how easily visual information is structured and navigated by the eye-brain system \parencite{reber2004processing}: a compelling image that effectively guide eye gaze can reduce visual cognitive load substantially.

Building upon these theories, we raise the question that \emph{whether it is both beneficial and feasible to incorporate human-like visual cognition into machine perception for AQA}. Models that overlook such a cognitive mechanism may risk semantic bias, overemphasizing semantic recognition (e.g., ``sunset,'' ``dog'') while lacking an intrinsic understanding of how visual flow shapes perception. As illustrated in Figure \ref{fig:gradcam}, pure semantic encoders (e.g., CLIP \parencite{radford2021learning} ) exhibit a rigid object-centric preference. By contrast, a human-like cognition encoder modeled on eye behavior scans composite scene properties such as foreground/background, chromatic stratification, luminance, illumination and inconsistencies, all of which significantly influence aesthetic judgment in addition to semantics.  

Inspired by the brain's Dual-Process Theory \parencite{kahneman2011thinking}, we propose \emph{AestheticNet}, a novel, cognitively inspired AQA paradigm that unites human-like visual cognition with semantic perception via a two-branch architecture, bridging the gap between ``recognition'' and ``appreciation'' of scenes. Aesthetic judgments emerge from the synergy between a dynamic, objective visual attention pathway and a static, subjective semantic interpretation pathway.

Concretely, AestheticNet augments semantic representations with HVS-derived visual attention captured by a dedicated gaze encoder, ensuring that aesthetic judgments reflect both human-viewing experience and machine-perceived semantics. This visual pathway utilizes a gaze-aligned visual encoder (GAVE), which is offline-pretrained on eye-tracking data to model how humans focus during aesthetic evaluation. We introduce Contrastive Gaze Alignment for pretraining, enabling resource-efficient learning that captures universal visual attentions with only about a hundred training images. The GAVE is incorporated as a complementary branch to the main semantic branch (e.g., a fixed CLIP encoder) \parencite{radford2021learning} via cross-attention fusion, offering visual cognitive priors to enhance AQA. Notably, the human-like visual cognition module serves as an orthogonal cognitive complement, computationally decoupled from semantic perception. Hence, HVS-derived attention functions as a model-agnostic corrector, compatible with diverse AQA backbones as a plug-and-play cognitive prior that consistently improves performance.

To rigorously validate the effect of human-like visual cognition in AQA, we conduct a hypothesis test on its necessity and modularity. Across extensive experiments with AestheticNet, we provide empirical evidence supporting the hypotheses that incorporating HVS-derived attention generally enhances AQA to significantly outperform semantic-only baselines. Moreover, these learned gaze attentions function as a flexible, model-agnostic corrector, showing seamless compatibility with diverse AQA backbones, from CNNs to MLLMs, as a plug-and-play biological prior module that consistently yields performance gains.


\section{Related Work}
\subsection{Computational Aesthetics}
The development of AQA marks a shift from rules-based modeling to deep representation learning. Early methods used handcrafted features, such as the rule of thirds, color harmony, and depth of field to quantify photographic rules \parencite{datta2006studying}. But these features, while explainable, failed to capture high-level semantic context. More advanced methods predict the score distribution directly from the global image to match the human-rating ground truth \parencite{talebi2018nima, su2020blindly}. Recently, multimodal large language models such as Q-Align have utilized massive text-image pre-training to align visual features with human aesthetic feedback \parencite{wu2024qalign}. However, these end-to-end deep representation methods treat aesthetics as static, image-centric attributes and may introduce semantic bias by ignoring the cognitive dynamics of how humans perceive visual aesthetics.

\subsection{Visual Attention in Aesthetics}
Visual attention modeling has predominantly focused on generic Saliency Detection \parencite{itti1998model}, which predicts fixation probability based on low-level feature contrast. However, aesthetic attention differs fundamentally from generic saliency; it is driven by artistic composition and visual balance rather than mere object distinctiveness. While prior works \parencite{ma2017lamp, lu2015deep} have integrated saliency maps derived from salient object detection paradigms into AQA pipelines, they inherently introduce an object-centric bias and fail to reflect the exploratory nature of human aesthetic viewing behaviors. Differently, AestheticNet encodes human visual system-derived attention via a dedicated Gaze Encoder. This mechanism internalizes the oculomotor priors associated with aesthetic appreciation, enabling a more precise assessment that transcends simple object recognition.

\begin{figure*}[t!]
\centering
\includegraphics[width=\textwidth]{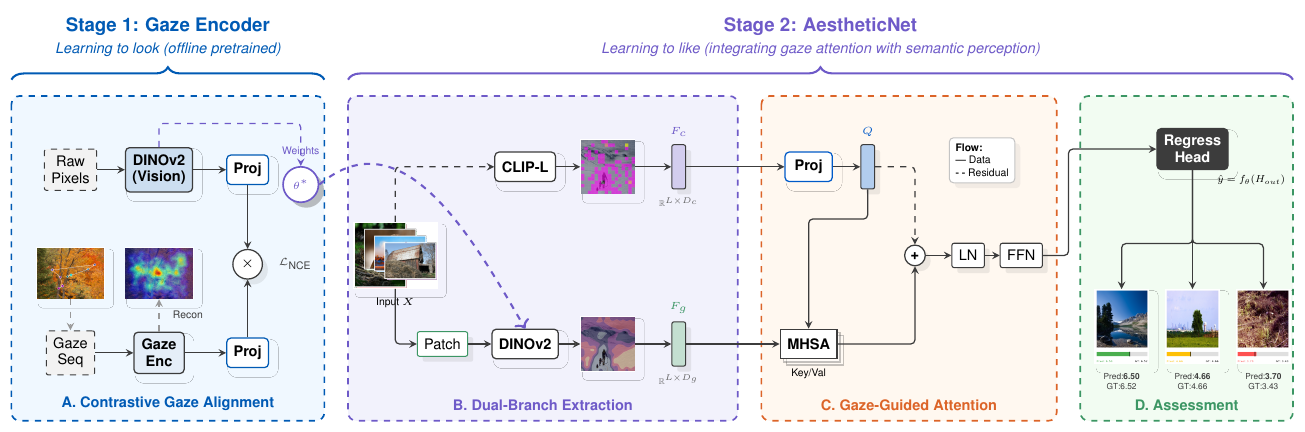}
\caption{\textbf{The Cognitive Architecture of AestheticNet.} This framework achieves aesthetic perception through two stages and four steps. \textbf{(A) Contrastive Gaze Alignment (CGA):} The Gaze Encoder aligns raw pixels with eye-tracking sequences using contrastive loss ($\mathcal{L}_{\text{CGA}}$) to learn a general "gaze grammar". \textbf{(B) Dual-Branch Extraction:} A frozen Semantic Encoder and the Gaze-Aligned Visual Encoder (GAVE)—the visual backbone of the pre-trained Gaze Encoder—extract content and perceptual form. \textbf{(C) Gaze-Guided Attention:} Semantic representations ($Q$) actively query gaze-aligned features ($K, V$) via a directed attention mechanism. \textbf{(D) Assessment:} The synthesized representation is mapped to a scalar aesthetic score $\hat{y}$. (MHSA: Multi-Head Self-Attention; LN: Layer Normalization; FFN: Feed-Forward Network).}
\label{fig:framework}
\end{figure*}


\section{Methods}

Our computational framework (visualized in Figure \ref{fig:framework}) operationalizes the cognitive interaction between active vision mechanisms and semantic interpretation. The architecture centers on the AestheticNet, which integrates a pre-trained, resource-rational Gaze Encoder to synthesize perceptual structure with semantic intent.

\subsection{The Gaze Encoder: Learning to Look}

Standard saliency models typically rely on supervised pixel-wise regression trained on large-scale datasets. We argue that this "big data" paradigm lacks ecological validity. In human development, visual attention is an early-emerging capability; infants demonstrate sophisticated scanpath patterns driven by low-level features months before acquiring the semantic vocabulary to name objects. This suggests that the mechanism of visual attention is not a high-dimensional semantic function requiring millions of samples, but a low-dimensional biological prior. To model this, in the first stage, we propose a resource-rational learning strategy utilizing Contrastive Gaze Alignment (CGA).

We utilize DINOv2 \parencite{oquab2023dinov2} (ViT-Small/14) as the computational proxy for the primary visual cortex (V1/V2). Unlike supervised CNNs that learn via explicit human annotation—analogous to formal instruction—DINOv2 \parencite{oquab2023dinov2}  derives representations solely from the statistical structure of visual data via self-supervision. Its self-attention mechanism naturally segments figure from ground without supervision, effectively providing a "phylogenetic" prior: a pre-wired sensitivity to visual structure that exists in the biological system prior to specific aesthetic training.

To fine-tune this biological backbone, we depart from data-hungry paradigms to reflect HVS resource rationality: evolving oculomotor priors minimizes the cognitive load of decoding every pixel semantically. We utilized a minimalist set of 109 images \parencite{wilming2017extensive}. Although the image count is small, they provide extremely dense, high-fidelity behavioral scanpaths. By fitting universal human sequential behaviors rather than memorizing static pixels, the model effectively prevents overfitting. Consequently, the Image Encoder distills the universal grammar of looking—such as saccadic amplitude and center bias—robustly from these dense traces.

We employ a dual-tower architecture where the Image Encoder ($E_I$, initialized with DINOv2 \parencite{oquab2023dinov2} ) and a Gaze Transformer ($E_G$) are aligned via a symmetric InfoNCE loss. This maximizes the mutual information between the visual representation and the aggregate human eye movement data. Let $s_{i,j} = \text{sim}(v_i^I, v_j^G)$ denote the similarity between the $i$-th image and $j$-th gaze embedding. The loss is defined as:

\begin{equation}
\mathcal{L}_{CGA} = - \frac{1}{2N} \sum_{i=1}^{N} \left[ \log \frac{e^{s_{i,i}/\tau}}{\sum_{k} e^{s_{i,k}/\tau}} + \log \frac{e^{s_{i,i}/\tau}}{\sum_{k} e^{s_{k,i}/\tau}} \right]
\end{equation}

where $\tau=0.05$ is the temperature parameter controlling the sharpness of the alignment. This objective forces the visual encoder to distinguish the specific oculomotor signature of an image from distractors, effectively "grounding" the DINOv2 \parencite{oquab2023dinov2}  features in biological behavior.

\subsection{AestheticNet: Learning to Like}

In the second stage, we construct the AestheticNet by synthesizing the learned gaze priors with high-level semantic understanding using a dual-pathway model. The model architecture consists of two parallel pathways: the Visual Attention Branch, which utilizes a pre-trained GAVE to extract perceptual features $h_F \in \mathbb{R}^{d}$ representing human vision dynamics; and the Semantic Branch, which employs a frozen CLIP-ViT-L/14 \parencite{radford2021learning} backbone to extract semantic embeddings $h_C \in \mathbb{R}^{d}$.

\textit{Gaze-Guided Attention Integration.}
Unlike passive concatenation strategies that treat form and content as static vectors, we model aesthetic judgment as an active interrogation process. We suggest that abstract semantic intents modulate the readout of perceptual features—a mechanism akin to top-down cognitive control. To operationalize this, we employ a Cross-Attention mechanism where the semantic latent vector governs the attentional focus over the perceptual stream. Formally, we define the semantic embedding as the Query ($Q$) and the perceptual features as both Key ($K$) and Value ($V$). The modulated representation and final prediction are computed as:

\begin{equation}
\begin{aligned}
h_{attn} &= \text{Softmax}\left(\frac{(h_C W_Q)(h_F W_K)^T}{\sqrt{d_k}}\right) (h_F W_V) \\
\hat{y} &= \mathcal{M}_{\phi}\left( \text{Concat}[h_{attn}, h_C] \right)
\end{aligned}
\end{equation}

where $W_Q, W_K, W_V$ are learnable projection matrices, $d_k$ is the scaling factor, and $\mathcal{M}_{\phi}$ denotes the final regression Multi-Layer Perceptron. This formulation mathematically simulates active inference: the projection $(h_C W_Q)$ constructs a "cognitive search template" that re-weights the structural features $(h_F)$ based on their relevance to the current semantic context, effectively filtering the "grammars of looking" through the lens of meaning.

\textit{Hybrid Optimization Objective.}
Human aesthetic judgment is characterized by a duality of absolute valuation (e.g., precise scoring) and ordinal ranking (e.g., preference sorting). To capture this, the network is optimized using a hybrid objective function. We combine Mean Squared Error (MSE) for numerical precision and a Pearson Linear Correlation Coefficient (PLCC) penalty for ranking linearity:

\begin{equation}
\begin{split}
\mathcal{L}_{total} &= \frac{1}{N}\sum_{i=1}^{N}(\hat{y}_i - y_i)^2 \\
&+ \lambda \left( 1 - \frac{\sum_{i}(\hat{y}_i - \mu_{\hat{y}})(y_i - \mu_y)}{\sqrt{\sum_{i}(\hat{y}_i - \mu_{\hat{y}})^2}\sqrt{\sum_{i}(y_i - \mu_y)^2}} \right)
\end{split}
\end{equation}

where $y_i$ and $\hat{y}_i$ represent the ground-truth and predicted scores, $\mu$ denotes the batch mean, and $\lambda=0.5$ is a balancing hyperparameter. The inclusion of the differentiable PLCC term (second line) constrains the model to learn the monotonic relationship of human preference, mitigating the regression-to-the-mean effect often observed in pure MSE optimization.


\section{Experiments}
To rigorously validate the effect of human visual cognition in AQA and challenge the prevailing semantic-along perspective, we devise the following hypothesis test with two studies: Study 1 evaluates the necessity of the visual attention pathway ($H_{1a}$), and Study 2 assesses its independence and potential as a plug-and-play module across diverse AQA backbones ($H_{1b}$).

\begin{hypothesis}
    \begin{description}
    \item[$H_0$ (The Null Hypothesis: Semantic Sufficiency):] 
    AQA primarily depends on semantic content; incorporating HVS-derived attention does not significantly improve performance over semantic-only baselines ($p > 0.05$).

    \item[$H_{1a}$ (The Necessity Hypothesis):] 
    Incorporating HVS-derived attention generally enhances AQA, as evidenced by AestheticNet, which fuses eye gaze attention with semantic representation, significantly outperforming semantic-only baselines ($p < 0.05$).

    \item[$H_{1b}$ (The Independence Hypothesis):] 
    Human-like visual cognition serves as an orthogonal cognitive complement, computationally decoupled from semantic perception. Hence, HVS-derived attention functions as a model-agnostic corrector, evidenced by seamless compatibility with diverse AQA backbones as a plug-and-play module that consistently yields performance gains.
\end{description}
\end{hypothesis}
We formally reject $H_0$ only if both $H_{1a}$ and $H_{1b}$ are verified by empirical evidence.

\subsection{Study 1: The Necessity of Human Visual Attention}
This study investigates whether integrating GAVE with the Semantic Encoder yields a statistically significant improvement in AQA compared to semantic-only baselines.

\paragraph{Dataset Construction.}
We evaluated our framework on a category-aligned subset of the AVA dataset \parencite{murray2012ava}. To ensure cognitive consistency with the Contrastive Gaze Alignment phase (which utilized compositional categories), we filtered the AVA dataset for 8 relevant tags (e.g., ``Landscape'', ``Architecture'', ``Still Life''). This resulted in a total of 89,677 images, split into 71,741 for training and 17,936 for testing.

\paragraph{Experimental Setup.}
The AestheticNet was instantiated with a frozen CLIP-ViT-L/14 \parencite{radford2021learning} (Semantic Encoder) and a DINOv2 \parencite{oquab2023dinov2}-based GAVE ($\sim$21M trainable parameters). These streams were integrated via Gaze-Guided Attention. The model was optimized on the hybrid loss (MSE and PLCC constraints) using AdamW (batch size 64, cosine learning rate scheduling). To prevent overfitting, training was bounded by a 200-epoch maximum and a 50-epoch early stopping criterion based on validation PLCC.

\subsubsection{Comparative Evaluation}
Having validated the necessity of the active vision mechanism, we benchmark AestheticNet against representative baselines: ResNet-50 \parencite{he2016deep}, NIMA \parencite{talebi2018nima}, HyperIQA \parencite{su2020blindly}, and CLIP-L \parencite{radford2021learning}. We also include Q-Align \parencite{wu2024qalign}, but designate it as a \emph{machine perception oracle} rather than a normal baseline. This is because Q-Align was trained on the full AVA dataset which covers the test subset in this experiment, rendering its performance a result of data memorization.

\begin{figure}[!t]
\centering
\includegraphics[width=\columnwidth]{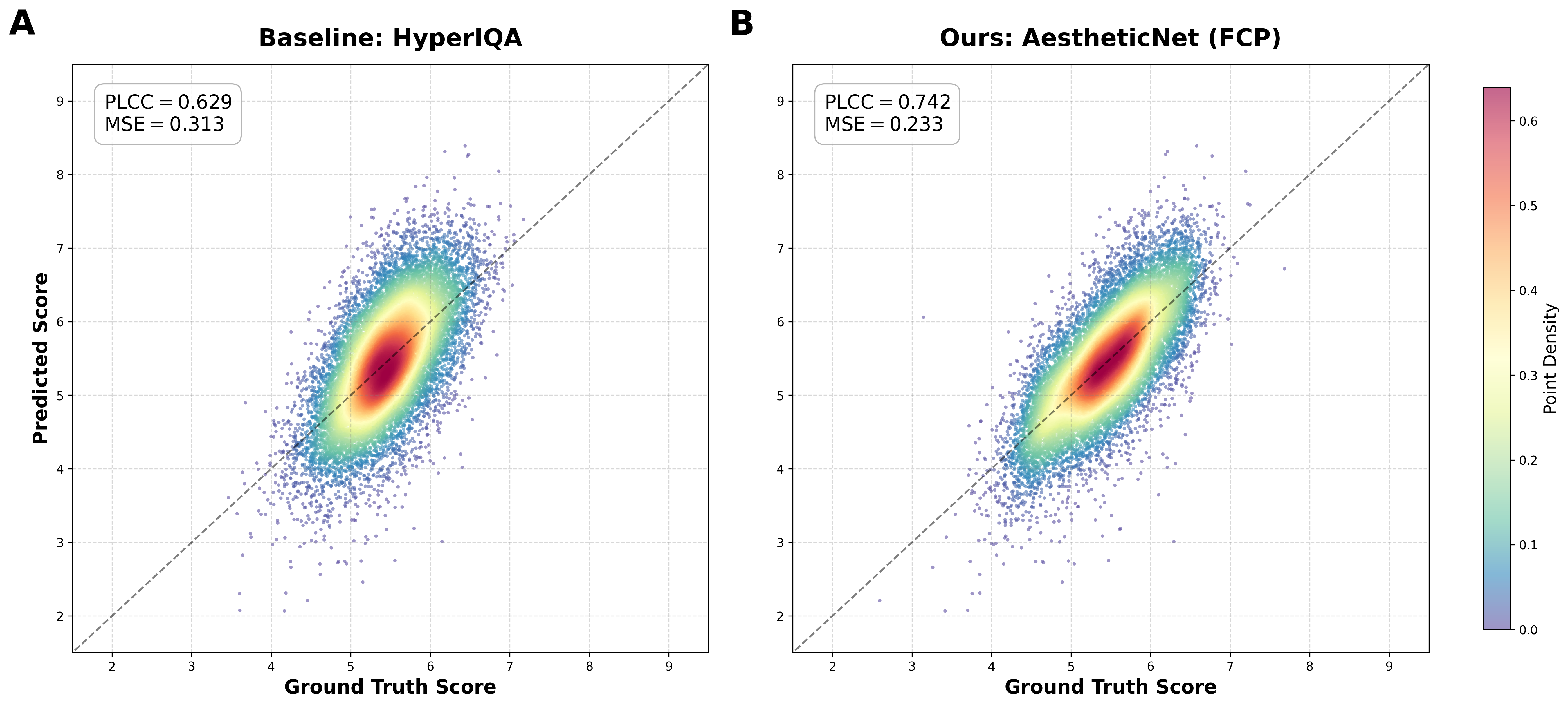}
\caption{\textbf{Prediction alignment analysis.} (A) The baseline HyperIQA shows scattered predictions with higher variance around the central diagonal. (B) AestheticNet produces a tighter distribution along the diagonal line ($y=x$), particularly in the high-density regions (red/yellow). This confirms that our dual-process approach aligns more closely with human consensus than single-stream baselines.}
\label{fig:scatter}
\end{figure}

As detailed in Table \ref{tab:sota}, AestheticNet (0.747) consistently outperforms all baseline architectures. Most critically, it surpasses its own semantic backbone, CLIP-L (0.617), by a substantial margin of $\Delta +0.130$. This empirical evidence directly supports $H_{1a}$, confirming that semantic representations alone are insufficient for accurate aesthetic assessment. Surpassing expert-level human benchmarks ($\text{SROCC}\approx0.711$ \parencite{kong2016photo}), AestheticNet captures collective consensus effectively. As Figure \ref{fig:scatter} visualizes, compared to HyperIQA—selected as the strongest traditional baseline avoiding massive multimodal confounders—our dual-process model yields a tighter diagonal distribution, visually confirming this superior alignment.

\begin{table}[!t]
\centering
\caption{\textbf{Baseline comparison.} AestheticNet significantly outperforms all baselines. All metrics include $95\%$ Confidence Intervals. * indicates $p<.001$ vs. AestheticNet.}
\label{tab:sota}
\scriptsize 
\setlength{\tabcolsep}{0.5pt} 
\begin{tabular*}{\columnwidth}{@{\extracolsep{\fill}}lccc}
\toprule
\textbf{Model} & \textbf{PLCC [95\% CI]} & \textbf{SROCC [95\% CI]} & \textbf{MSE [95\% CI]} \\
\midrule
\multicolumn{4}{l}{\textit{Dataset: Full Test Set}} \\
ResNet-50 & 0.559* [0.549, 0.568] & 0.548* [0.538, 0.559] & 0.359* [0.351, 0.368] \\
NIMA & 0.584* [0.574, 0.593] & 0.575* [0.565, 0.586] & 0.342* [0.334, 0.350] \\
CLIP-L & 0.617* [0.607, 0.625] & 0.618* [0.608, 0.628] & 0.392* [0.383, 0.401] \\
HyperIQA & 0.629* [0.619, 0.638] & 0.615* [0.604, 0.624] & 0.314* [0.306, 0.322] \\
\textbf{AestheticNet} & \textbf{0.747} [0.740, 0.754] & \textbf{0.740} [0.732, 0.747] & \textbf{0.261} [0.254, 0.267] \\
\midrule 
\textit{Q-Align (oracle)} & \textit{0.878} [0.874, 0.882] & \textit{0.875} [0.871, 0.879] & \textit{0.117} [0.114, 0.121] \\
\textit{Human (expert-level)} & & \textit{0.711} &  \\
\midrule
\multicolumn{4}{l}{\textit{Dataset: Top-10k Lowest Error Samples}} \\
CLIP-L & 0.655* [0.642, 0.666] & 0.624* [0.611, 0.637] & 0.067* [0.064, 0.069] \\
NIMA & 0.859* [0.853, 0.865] & 0.840* [0.832, 0.847] & 0.057* [0.055, 0.059] \\
ResNet-50 & 0.869* [0.863, 0.874] & 0.855* [0.848, 0.862] & 0.059* [0.057, 0.061] \\
HyperIQA & 0.913* [0.909, 0.916] & 0.899* [0.894, 0.904] & 0.050* [0.048, 0.052] \\
\textbf{AestheticNet} & \textbf{0.953} [0.951, 0.955] & \textbf{0.950} [0.947, 0.952] & \textbf{0.042} [0.041, 0.043] \\
\midrule
\textit{Q-Align (oracle)} & \textit{0.978} [0.977, 0.979] & \textit{0.969} [0.968, 0.971] & \textit{0.018} [0.017, 0.018] \\
\bottomrule
\end{tabular*}
\end{table}



\subsubsection{Category-wise Analysis}

To verify that the gaze-guided improvement is robust across diverse semantic domains and not driven by a single class, we decomposed the performance by 8 distinct categories defined in the AVA dataset.

\begin{table}[!t]
\centering
\caption{\textbf{Category-wise performance.} Evaluated on the full test set, decomposed by semantic category. AestheticNet consistently outperforms the semantic baseline (S-Only) across all 8 categories. The improvement is particularly pronounced in structure-heavy domains like Still Life and Architecture. * indicates $p < .001$ compared to AestheticNet.}
\label{tab:category}
\scriptsize
\setlength{\tabcolsep}{1.5pt}{
\begin{tabular*}{\columnwidth}{@{\extracolsep{\fill}}lccc}
\toprule
\textbf{Category (N)} & \textbf{Model} & \textbf{PLCC [95\% CI]} & \textbf{MSE [95\% CI]} \\
\midrule
\textbf{Nature} & S-Only & 0.664* [0.647, 0.681] & 0.348* [0.332, 0.365] \\
(N=4464) & \textbf{AestheticNet} & \textbf{0.737} [0.722, 0.752] & \textbf{0.264} [0.250, 0.278] \\
\midrule
\textbf{Landscape} & S-Only & 0.711* [0.695, 0.725] & 0.332* [0.317, 0.348] \\
(N=4062) & \textbf{AestheticNet} & \textbf{0.763} [0.750, 0.776] & \textbf{0.256} [0.244, 0.268] \\
\midrule
\textbf{Cityscape} & S-Only & 0.715* [0.690, 0.739] & 0.332* [0.307, 0.358] \\
(N=1408) & \textbf{AestheticNet} & \textbf{0.751} [0.729, 0.772] & \textbf{0.271} [0.250, 0.295] \\
\midrule
\textbf{Architecture} & S-Only & 0.655* [0.625, 0.683] & 0.304* [0.281, 0.328] \\
(N=1735) & \textbf{AestheticNet} & \textbf{0.732} [0.708, 0.752] & \textbf{0.233} [0.216, 0.251] \\
\midrule
\textbf{Still Life} & S-Only & 0.632* [0.611, 0.653] & 0.359* [0.341, 0.378] \\
(N=3752) & \textbf{AestheticNet} & \textbf{0.719} [0.701, 0.737] & \textbf{0.265} [0.250, 0.282] \\
\midrule
\textbf{Water} & S-Only & 0.673* [0.630, 0.711] & 0.396* [0.352, 0.448] \\
(N=732) & \textbf{AestheticNet} & \textbf{0.755} [0.724, 0.787] & \textbf{0.277} [0.247, 0.310] \\
\midrule
\textbf{Sky} & S-Only & 0.716* [0.685, 0.746] & 0.335* [0.301, 0.365] \\
(N=1102) & \textbf{AestheticNet} & \textbf{0.758} [0.731, 0.784] & \textbf{0.297} [0.270, 0.326] \\
\midrule
\textbf{Rural} & S-Only & 0.700* [0.661, 0.735] & 0.296* [0.264, 0.331] \\
(N=681) & \textbf{AestheticNet} & \textbf{0.749} [0.715, 0.781] & \textbf{0.228} [0.204, 0.255] \\
\bottomrule
\end{tabular*}}
\end{table}

As shown in Table \ref{tab:category}, AestheticNet yields statistically significant improvements ($p < .001$) across all 8 semantic categories, confirming that the benefit of the active vision pathway is universal rather than niche. 

Notably, the performance gains are non-uniform. The most substantial improvements are observed in Still Life ($\Delta +0.087$) and Architecture ($\Delta +0.077$). These domains are inherently driven by geometric composition, spatial arrangement, and visual balance—features that are explicitly encoded by the GAVE but often under-represented in semantic models (S-Only). For instance, in Still Life, the aesthetic value often derives from the specific framing of objects rather than the objects themselves. By capturing these compositional priors, AestheticNet effectively compensates for the structural blindness of the pure semantic backbone.


\subsubsection{Ablation Study}

To isolate the contribution of the active vision pathway and test the Necessity Hypothesis ($H_{1a}$), we performed a component-wise ablation on the fully trained AestheticNet. Instead of training separate models, we utilized inference-time masking to measure the reliance of the joint architecture on each stream, following three configurations:

\begin{itemize}
    \item \textit{G-Only:} The Semantic stream is masked (zeroed out) during inference, forcing the model to rely solely on the GAVE.
    \item \textit{S-Only:} The Gaze stream is masked during inference, relying solely on the frozen Semantic Encoder ($H_0$ baseline).
    \item \textit{AestheticNet:} The full architecture with both streams active.
\end{itemize}

\begin{table}[ht]
\centering
\caption{\textbf{Ablation results.} Comparison of isolated branches (G-Only, S-Only) versus the complete model via inference-time masking. All metrics include 95\% Confidence Intervals in brackets. * indicates $p < .001$ compared to AestheticNet.}
\label{tab:ablation}
\scriptsize 
\setlength{\tabcolsep}{1.5pt} 
\begin{tabular*}{\columnwidth}{@{\extracolsep{\fill}}lccc}
\toprule
\textbf{Model} & \textbf{PLCC [95\% CI]} & \textbf{SROCC [95\% CI]} & \textbf{MSE [95\% CI]} \\
\midrule
\multicolumn{4}{l}{\textit{Dataset: Full Test Set}} \\
G-Only & 0.607* [0.598, 0.617] & 0.596* [0.586, 0.606] & 0.415* [0.405, 0.424] \\
S-Only & 0.682* [0.674, 0.691] & 0.678* [0.669, 0.687] & 0.340* [0.333, 0.348] \\
\textbf{AestheticNet} & \textbf{0.746} [0.740, 0.753] & \textbf{0.740} [0.732, 0.747] & \textbf{0.261} [0.255, 0.268] \\
\midrule
\multicolumn{4}{l}{\textit{Dataset: Top-10k Lowest Error Samples}} \\
G-Only & 0.912* [0.908, 0.916] & 0.892* [0.887, 0.897] & 0.069* [0.067, 0.071] \\
S-Only & 0.941* [0.938, 0.943] & 0.939* [0.936, 0.942] & 0.056* [0.054, 0.058] \\
\textbf{AestheticNet} & \textbf{0.953} [0.951, 0.955] & \textbf{0.950} [0.947, 0.952] & \textbf{0.042} [0.040, 0.043] \\
\bottomrule
\end{tabular*}
\end{table}

Table \ref{tab:ablation} presents the comparison. On the full test set, S-Only yields a PLCC of 0.682. Because our cross-attention employs residual connections, masking GAVE downgrades the network to a stable semantic regressor, averting out-of-distribution collapse. Re-activating both pathways (AestheticNet) significantly boosts performance to 0.746, confirming the GAVE provides indispensable structural information. The analysis of the \textit{Top-10k Lowest Error Samples} further validates this dependency. Even in this high-precision regime, masking either branch (G-Only or S-Only) degrades performance compared to the full model (0.953). This implies that "peak aesthetic judgment" is functionally irreducible: it requires the simultaneous integration of visual flow and semantic content.





\subsection{Study 2: The Independence of Human Visual Attention}

In this study, we test the Independence Hypothesis ($H_{1b}$): if the GAVE captures an orthogonal biological signal, it should function as a universal cognitive patch, improving performance regardless of the host architecture.

\paragraph{Plug-and-Play Integration.}
We integrated the frozen GAVE into four diverse backbones using two strategies adapted to the host's accessibility:
\begin{itemize}
    \item \textit{Feature-Level Fusion:} For standard backbones ResNet-50 \parencite{he2016deep}, NIMA \parencite{talebi2018nima}, HyperIQA \parencite{su2020blindly}, CLIP-L \parencite{radford2021learning}, we concatenated the gaze features directly with the host's penultimate feature vector ($f_{fused} = f_{host} \oplus f_{gaze}$) before the final regression head.
    \item \textit{Score-Level Correction:} For the large-scale Q-Align \parencite{wu2024qalign}, we adopted a non-intrusive strategy by applying a lightweight residual correction to the final predicted scores ($S_{final} = S_{host} + \lambda \cdot S_{gaze}$).
\end{itemize}

\begin{table}[!t]
\centering
\caption{\textbf{Plug-and-play performance} evaluated on the full test set. Integrating the gaze attention yields statistically significant improvements across all architectures. * indicates $p < .001$, ** indicates $p < .01$ (Original vs. +Gaze).}
\label{tab:plug_and_play}
\scriptsize 
\setlength{\tabcolsep}{1.5pt} 
\begin{tabular*}{\columnwidth}{@{\extracolsep{\fill}}lccc}
\toprule
\textbf{Model / Config} & \textbf{PLCC [95\% CI]} & \textbf{SROCC [95\% CI]} & \textbf{MSE [95\% CI]} \\
\midrule
\multicolumn{4}{l}{\textbf{ResNet-50}} \\
Original & 0.559* [0.549, 0.568] & 0.548* [0.538, 0.559] & 0.359* [0.351, 0.368] \\
+ Gaze & \textbf{0.635} [0.625, 0.644] & \textbf{0.624} [0.614, 0.635] & \textbf{0.321} [0.313, 0.330] \\
\midrule
\multicolumn{4}{l}{\textbf{CLIP-L}} \\
Original & 0.617* [0.607, 0.625] & 0.618* [0.608, 0.628] & 0.392* [0.383, 0.401] \\
+ Gaze & \textbf{0.737} [0.728, 0.746] & \textbf{0.730} [0.719, 0.739] & \textbf{0.242} [0.233, 0.251] \\
\midrule
\multicolumn{4}{l}{\textbf{NIMA}} \\
Original & 0.584* [0.574, 0.593] & 0.575* [0.565, 0.586] & 0.342* [0.334, 0.350] \\
+ Gaze & \textbf{0.625} [0.615, 0.634] & \textbf{0.616} [0.605, 0.626] & \textbf{0.330} [0.322, 0.338] \\
\midrule
\multicolumn{4}{l}{\textbf{HyperIQA}} \\
Original & 0.629* [0.619, 0.638] & 0.615* [0.604, 0.624] & 0.314* [0.306, 0.322] \\
+ Gaze & \textbf{0.684} [0.674, 0.693] & \textbf{0.675} [0.664, 0.684] & \textbf{0.277} [0.269, 0.285] \\
\midrule
\multicolumn{4}{l}{\textit{\textbf{Q-Align (13B)}}} \\
\textit{Original} & \textit{0.878}* [0.874, 0.882] & \textit{0.875}** [0.871, 0.879] & \textit{0.117}* [0.114, 0.121] \\
\textit{+ Gaze} & \textbf{\textit{0.879}} [0.875, 0.883] & \textbf{\textit{0.876}} [0.871, 0.879] & \textbf{\textit{0.116}} [0.113, 0.120] \\
\bottomrule
\end{tabular*}
\end{table}

Table \ref{tab:plug_and_play} demonstrates that incorporating gaze prior information significantly improves the performance of all evaluation baselines. The magnitude of the performance improvement is related to the host model: the CLIP-L model \parencite{radford2021learning}, which heavily relies on semantic content, benefits the most ($\Delta \text{PLCC} +0.120$), achieving a huge improvement simply by adding the GAVE.

More broadly, the universality of these gains across architecturally distinct baselines—from standard CNNs to specialized aesthetic regressors—supports a deeper conclusion. Despite their structural differences, these models all lack explicit oculomotor modeling. The GAVE functions as a model-agnostic cognitive corrector, compensating for the lack of active viewing dynamics in static deep learning models. This consistent enhancement verifies the Independence Hypothesis ($H_{1b}$), suggesting that the ``grammar of looking'' operates as an orthogonal cognitive prior computationally decoupled from visual feature encoding. Finally, regarding Q-Align \parencite{wu2024qalign}, which serves as a \emph{machine perception oracle} in this context, we still observe a statistically significant refinement ($p < .001$). This confirms that the instinctual scanpath signatures captured by our module remain an orthogonal and irreducible component, indispensable even at the ceiling of machine perception.



\section{Discussion}

Our results formally reject the Null Hypothesis of Semantic Sufficiency ($H_0$). We validate the Necessity Hypothesis ($H_{1a}$) through three converging lines of evidence: comparative evaluations show AestheticNet significantly outperforms semantic-only baselines; category-wise analysis reveals robust gains across diverse compositional domains; and ablation studies verify the functional necessity of the Gaze Branch. Additionally, we confirm the Independence Hypothesis ($H_{1b}$), as the GAVE functions as a seamless, model-agnostic corrector across diverse architectures. Together, these findings prove that aesthetic judgment is not just a passive semantic readout, but an active integration process where oculomotor dynamics modulate semantic interpretation.

Although biological gaze intertwines diverse signals, our plug-and-play experiments demonstrate a clear computational decoupling. The GAVE's ability to enhance architectures ranging from CNNs to Q-Align \parencite{wu2024qalign} confirms that the ``grammar of looking'' acts as an orthogonal cognitive complement. This suggests that robust aesthetic AI can be built by assembling specialized cognitive modules rather than relying solely on monolithic models, offering a path toward more interpretable systems.

However, our conclusions are bounded by current benchmarks. The AVA dataset follows a Gaussian distribution, creating data sparsity at aesthetic extremes. While AestheticNet handles these regions better than baselines, it remains constrained by this central tendency bias. Furthermore, the reliance on photography limits generalization to non-photographic domains, such as abstract art, where biological scanning strategies may differ.

\section{Conclusion}

This study establishes a resource-rational framework for computational aesthetics, effectively bridging active vision mechanisms with semantic interpretation. By operationalizing the synergy between the GAVE and the Semantic Encoder, we demonstrate that aesthetic perception relies not merely on high-dimensional feature mapping, but on the dynamic integration of innate oculomotor priors with semantic content. Our findings confirm that the GAVE functions as an independent, plug-and-play cognitive module, enabling significant performance gains with high data efficiency. Ultimately, AestheticNet advocates for a shift from monolithic modeling to biologically grounded, modular architectures. 

\section{Acknowledgement}
This work was supported by the National Natural Science Foundation of China under Grant No. 62402009, the Key Laboratory of Computing Power Network and Information Security, Ministry of Education under Grant No.2024PY014, and the Science and Technology Development Fund of Macao under Grant No. 0013-2024-ITP1 and 0069/2025/ITP2.

\printbibliography

@book{shimamura2012aesthetic,
  editor    = {Shimamura, Arthur P. and Palmer, Stephen E.},
  title     = {Aesthetic science: Connecting minds, brains, and experience},
  publisher = {Oxford University Press},
  address   = {New York, NY},
  date      = {2012},
}

@inproceedings{murray2012ava,
  author    = {Murray, Naila and Marchesotti, Luca and Perronnin, Florent},
  title     = {{AVA}: A large-scale database for aesthetic visual analysis},
  booktitle = {2012 IEEE Conference on Computer Vision and Pattern Recognition ({CVPR})},
  date      = {2012},
  pages     = {2408--2415},
  publisher = {{IEEE}},
  doi       = {10.1109/CVPR.2012.6247954}
}

@article{talebi2018nima,
  author       = {Talebi, Hossein and Milanfar, Peyman},
  title        = {{NIMA}: Neural image assessment},
  journaltitle = {{IEEE} Transactions on Image Processing},
  volume       = {27},
  number       = {8},
  date         = {2018},
  pages        = {3998--4011},
  doi          = {10.1109/TIP.2018.2831899}
}

@inproceedings{radford2021learning,
  author    = {Radford, Alec and Kim, Jong Wook and Hallacy, Chris and Ramesh, Aditya and Goh, Gabriel and Agarwal, Sandhini and Sastry, Girish and Askell, Amanda and Mishkin, Pamela and Clark, Jack and others},
  title     = {Learning transferable visual models from natural language supervision},
  booktitle = {International Conference on Machine Learning ({ICML})},
  series    = {Proceedings of Machine Learning Research},
  volume    = {139},
  date      = {2021},
  pages     = {8748--8763},
  publisher = {{PMLR}}
}

@article{noton1971scanpaths,
  author       = {Noton, David and Stark, Lawrence},
  title        = {Scanpaths in eye movements during pattern perception},
  journaltitle = {Science},
  volume       = {171},
  number       = {3968},
  date         = {1971},
  pages        = {308--311},
  doi          = {10.1126/science.171.3968.308}
}

@article{wilming2017extensive,
  author       = {Wilming, Niklas and Onat, Selim and Ossand{\'o}n, Jos{\'e} P. and A{\c{c}}{\i}k, Alper and Kietzmann, Tim C. and Kaspar, Kai and Gameiro, Ricardo R. and Vormberg, Alexandra and K{\"o}nig, Peter},
  title        = {An extensive dataset of eye movements during viewing of complex images},
  journaltitle = {Scientific Data},
  volume       = {4},
  number       = {1},
  date         = {2017},
  pages        = {160126},
  doi          = {10.1038/sdata.2016.126}
}

@article{oquab2023dinov2,
  author       = {Oquab, Maxime and Darcet, Timoth{\'e}e and Moutakanni, Th{\'e}o and Vo, Huy and Szafraniec, Marc and Khalidov, Vasil and Fernandez, Pierre and Haziza, Daniel and Massa, Francisco and El-Nouby, Alaaeldin and others},
  title        = {{DINOv2}: Learning robust visual features without supervision},
  journaltitle = {Transactions on Machine Learning Research},
  date         = {2024}
}

@inproceedings{wu2024qalign,
  author    = {Wu, Haoning and Zhang, Zicheng and Zhang, Weixia and Chen, Chaofeng and Liao, Liang and Li, Chunyi and Gao, Yixuan and Wang, Annan and Zhang, Erli and others},
  title     = {{Q-Align}: Teaching large visual-language models to align with human judgments},
  booktitle = {International Conference on Machine Learning ({ICML})},
  date      = {2024}
}

@inproceedings{su2020blindly,
  author    = {Su, Shaolin and Yan, Qingsen and Zhu, Yu and Zhang, Cheng and Ge, Xin and Sun, Jinqiu and Zhang, Yanning},
  title     = {Blindly assess image quality in the wild guided by a self-adaptive hyper network},
  booktitle = {2020 IEEE/CVF Conference on Computer Vision and Pattern Recognition ({CVPR})},
  date      = {2020},
  pages     = {3667--3676},
  publisher = {{IEEE}}
}

@book{yarbus1967eye,
  author    = {Yarbus, Alfred L.},
  title     = {Eye Movements and Vision},
  publisher = {Plenum Press},
  address   = {New York, NY},
  date      = {1967}
}

@article{reber2004processing,
  author       = {Reber, Rolf and Schwarz, Norbert and Winkielman, Piotr},
  title        = {Processing fluency and aesthetic pleasure: Is beauty in the perceiver's processing experience?},
  journaltitle = {Personality and Social Psychology Review},
  volume       = {8},
  number       = {4},
  date         = {2004},
  pages        = {364--382},
  doi          = {10.1207/s15327957pspr0804_3}
}

@article{itti1998model,
  author       = {Itti, Laurent and Koch, Christof and Niebur, Ernst},
  title        = {A model of saliency-based visual attention for rapid scene analysis},
  journaltitle = {{IEEE} Transactions on Pattern Analysis and Machine Intelligence},
  volume       = {20},
  number       = {11},
  date         = {1998},
  pages        = {1254--1259},
  doi          = {10.1109/34.730558}
}

@book{kahneman2011thinking,
  author    = {Kahneman, Daniel},
  title     = {Thinking, Fast and Slow},
  publisher = {Farrar, Straus and Giroux},
  address   = {New York, NY},
  date      = {2011}
}

@inproceedings{datta2006studying,
  author    = {Datta, Ritendra and Joshi, Dhiraj and Li, Jia and Wang, James Z.},
  title     = {Studying aesthetics in photographic images using a computational approach},
  booktitle = {European Conference on Computer Vision ({ECCV})},
  date      = {2006},
  pages     = {288--301},
  publisher = {Springer},
  doi       = {10.1007/11744078_23}
}

@inproceedings{kong2016photo,
  author    = {Kong, Shu and Shen, Xiaohui and Lin, Zhe and Mech, Radomir and Fowlkes, Charless},
  title     = {Photo aesthetics ranking network with attributes and content adaptation},
  booktitle = {European Conference on Computer Vision ({ECCV})},
  date      = {2016},
  pages     = {662--679},
  publisher = {Springer},
  doi       = {10.1007/978-3-319-46448-0_40}
}

@inproceedings{ma2017lamp,
  author    = {Ma, Shuang and Liu, Jing and Chen, Chang Wen},
  title     = {{A-Lamp}: Adaptive layout-aware multi-patch deep convolutional neural network for photo aesthetic assessment},
  booktitle = {2017 IEEE Conference on Computer Vision and Pattern Recognition ({CVPR})},
  date      = {2017},
  pages     = {4535--4544},
  publisher = {{IEEE}}
}

@inproceedings{lu2015deep,
  author    = {Lu, Xin and Lin, Zhe and Shen, Xiaohui and Mech, Radomir and Wang, James Z.},
  title     = {Deep multi-patch aggregation network for image style, aesthetics, and quality estimation},
  booktitle = {2015 IEEE International Conference on Computer Vision ({ICCV})},
  date      = {2015},
  pages     = {990--998},
  publisher = {{IEEE}},
  doi       = {10.1109/ICCV.2015.119}
}

@inproceedings{he2016deep,
  author    = {He, Kaiming and Zhang, Xiangyu and Ren, Shaoqing and Sun, Jian},
  title     = {Deep residual learning for image recognition},
  booktitle = {2016 IEEE Conference on Computer Vision and Pattern Recognition ({CVPR})},
  date      = {2016},
  pages     = {770--778},
  publisher = {{IEEE}},
  doi       = {10.1109/CVPR.2016.90}
}

\end{document}